\documentclass[journal]{IEEEtran}

\usepackage{times}
\usepackage{tabularx}
\usepackage{multirow}
\usepackage[dvipsnames]{xcolor}
\usepackage{booktabs}
\usepackage{balance}
\usepackage{graphicx}
\usepackage{comment}
\usepackage{amsmath}
\usepackage{amsthm}
\usepackage{amssymb}
\usepackage{amsfonts}
\usepackage{array}
\usepackage{url}
\usepackage{textcomp}
\usepackage[tight]{subfigure}
\usepackage{hhline}

\usepackage[normalem]{ulem}
\usepackage{ulem}
\usepackage{longtable}
\usepackage{rotating}
\usepackage{bm}
\usepackage{float}
\usepackage{flushend}
\usepackage{enumitem}
\usepackage{supertabular}
\usepackage{mathtools}
\usepackage{pifont}
\usepackage{cases}

\usepackage{algorithm}
\usepackage{algpseudocode}
\usepackage{amsmath}

\newcolumntype{C}[1]{>{\centering\let\newline\\\arraybackslash\hspace{0pt}}m{#1}}

\begin{document}
%

\title{Generative Partial Multi-View Clustering}
%
%
%

\author{Qianqian Wang,
        Zhengming Ding,
         Zhiqiang~Tao,
        Quanxue Gao,
        and Yun~Fu, \IEEEmembership{Fellow,~IEEE}
\thanks{Manuscript received Oct, 2020; revised ********; accepted ********. This work was partially done when Qianqian Wang was visiting Northeastern University as a visiting student.}
\thanks{Corresponding author: Q. Gao, (e-mail: qxgao@xidian.edu.cn).}
\thanks{Q. Wang, and Q. Gao are with the State Key Lab. Integrated Services Networks, Xidian University, 710071,  Xi'an China (email: qqwang@xidian.edu.cn).}
\thanks{Z. Ding is with the Department of Computer, Information and Technology, Indiana University-Purdue University Indianapolis, USA.}
\thanks{Z. Tao is with the Department of Electrical and Computer Engineering, Northeastern University, Boston, MA 02115 USA (email: zqtao@ece.neu.edu).}
\thanks{Y. Fu is with the Department of Electrical and Computer Engineering, Khoury College of Computer Sciences, Northeastern University, Boston, MA 02115 USA (email: yunfu@ece.neu.edu).}
}
%
%

\markboth{Journal of \LaTeX\ Class Files,~Vol.~14, No.~8, August~2020}%
{Shell \MakeLowercase{\textit{et al.}}: Bare Demo of IEEEtran.cls for IEEE Journals}
%



\maketitle

\begin{abstract}
Nowadays, with the rapid development of data collection sources and feature extraction methods, multi-view data are getting easy to obtain and have received increasing research attention in recent years, among which, multi-view clustering (MVC) forms a mainstream research direction and is widely used in data analysis. However, existing MVC methods mainly assume that each sample appears in all the views, without considering the incomplete view case due to data corruption, sensor failure, equipment malfunction, etc. In this study, we design and build a generative partial multi-view clustering model, named as GP-MVC, to address the incomplete multi-view problem by explicitly generating the data of missing views. The main idea of GP-MVC lies at two-fold. First, multi-view encoder networks are trained to learn common low-dimensional representations, followed by a clustering layer to capture the consistent cluster structure across multiple views.  Second, view-specific generative adversarial networks are developed to generate the missing data of one view conditioning on the shared representation given by other views. These two steps could be promoted mutually, where learning common representations facilitates data imputation and the generated data could further explores the view consistency. Moreover, an weighted adaptive fusion scheme is implemented to exploit the complementary information among different views. Experimental results on four benchmark datasets are provided to show the effectiveness of the proposed GP-MVC over the state-of-the-art methods.
\end{abstract}


%
\IEEEpeerreviewmaketitle

\section{Introduction}

\IEEEPARstart{M}{ulti-view} data could be samples collected from multiple sources, modalities captured by various sensors, or features extracted with different methods. Owing to the advance of hardware technology, multi-view data are quite common in real world~\cite{cc1}. For example, an image in social network could be represented by either its visual cues or the users' comments on it. Compared with single-view data, multiple views usually boost the model performance~\cite{Hou2017Discriminative,zhang2018generalized,Ding2018Robust} by providing the complementary information to represent the same data.


In recent years, increasing research efforts have been made in multi-view learning, where multi-view clustering (MVC)~\cite{xie2020adaptive,chao2017survey,tang2018learning,tao2019marginalized,zhan2018graph} forms a mainstream task that aims to explore the underlying data cluster structure shared by multiple views. MVC methods work as an effective data analysis tool for unlabeled multi-view datasets, and could significantly improve the clustering performance by fusing the information from different views. However, although traditional MVC methods achieve promising progress, their effectiveness depends on the completeness assumption for all the views of each instance. Hence, their performance may degrade when some views include missing data, which raise a challenging case as \emph{partial multi-view data}~\cite{cc25} or \emph{incomplete view data}~\cite{cc29,cc30}.


In practice, incomplete-view data are quite ubiquitous due to the environment issues, obstacle, noise, and malfunction of the collection/transmission/storage equipments~\cite{cc25}. For example, in medical data, some patients may not finish a complete examination for time conflict or other reason; in social multimedia, some instances may not contain visual or audio view as a result of the sensor breakdown. However, traditional MVC methods cannot handle incomplete challenges directly because they aim to find a shared structure among all views and require completeness of each data. In light of this, partial multi-view clustering (PMVC) algorithms have been developed~\cite{cc25,cc29,cc30,cc31,zhang2019cpm,liu2018late}.

\begin{figure*}[!t]
\centering
\includegraphics[width=16.5cm]{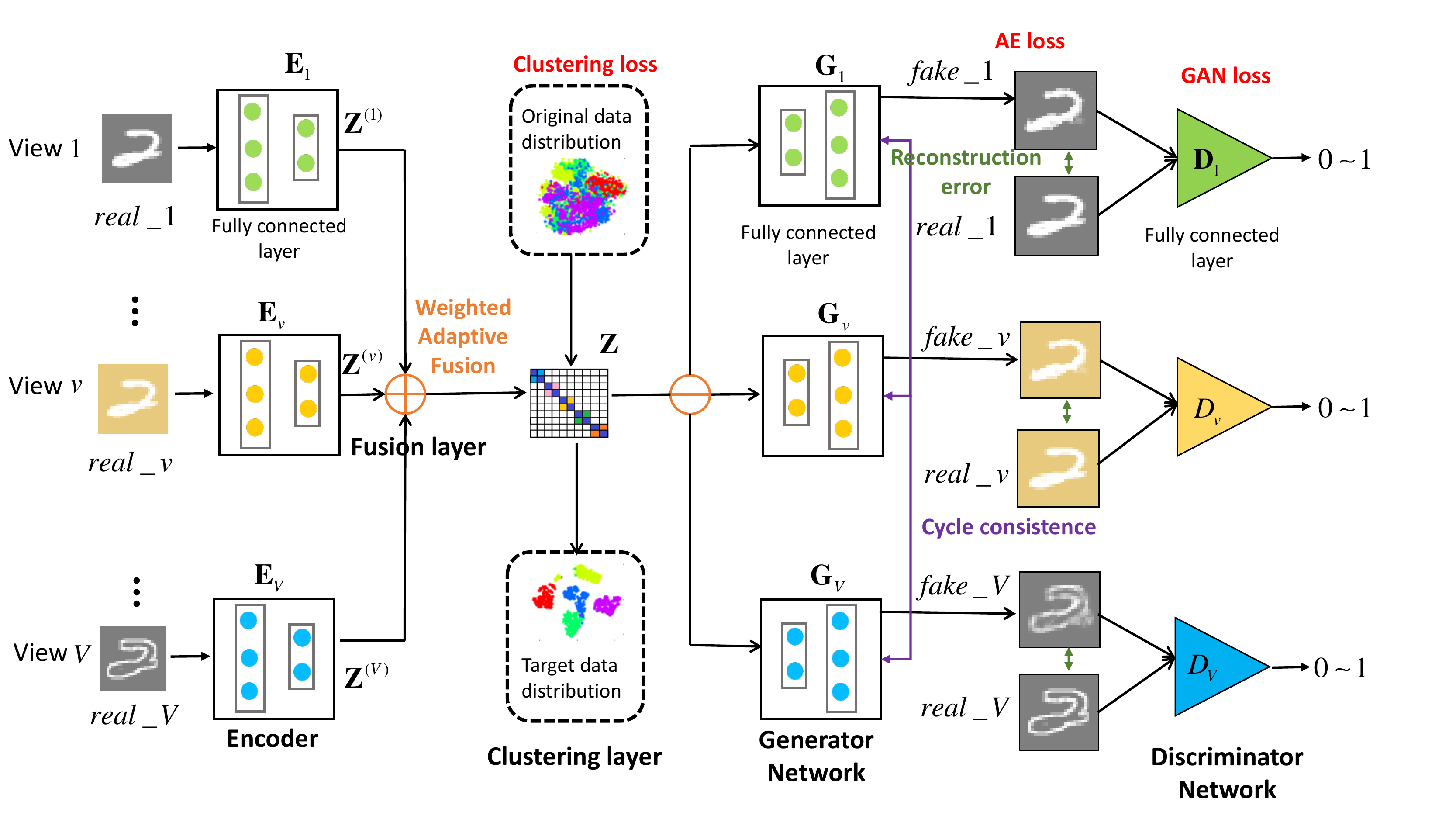}
 \vspace{-0.3cm}
\caption{The framework of our model. It consists of multi-view encoders ${\bf{E}}$, a weighted adaptive fusion layer, a deep embedding clustering layer, multi-view generators ${{\bf{G}}}$, and multi-view discriminators ${\bf{D}}$.}\label{fig1}
 \vspace{-0.3cm}
\end{figure*}

In the pioneering works, PMVC methods simply use zero or mean value to fill up the incomplete views. However, these simply imputed data are quite different from the real ones, which avoid MVC to learn a consistent clustering structure and badly degrade the final clustering performance. Existing PMVC methods target to establish a shared latent subspace with complete views and then compensate the latent representations for the missing data, which could be divided into two main directions. Specifically, the first kind is kernel based methods~\cite{cc29,liu2018late}, where the main idea is to leverage kernel matrices of the complete views for completing the kernel matrix of incomplete view. This kind of method can only be applied in kernel-based multi-view clustering. The second kind of methods are based on the non-negative matrix factorization (NMF)~\cite{cc26,cc27}. For sample missing from a certain view, these methods recover the non-negative matrix corresponding to the view with those obtained from the sample of which the view is un-missing. However, these two kinds of methods still have several limitations. (1) They need to process all the data together and it is inefficient to employ them for the large-scale databases. (2) They require numerous inverse operations for matrix factorization, resulting in a high time complexity. (3) They mainly exploit some regularization and add some constraints on the new representation, yet fail in compensating the missing data in each view explicitly. 

Inspired by generative adversarial networks (GAN)~\cite{zhang2019image,gui2020review,karras2019style},
it is natural to synthesize the missing data for learning representations of partial views. Vanilla GAN~\cite{cc38} was proposed to generate desired data from random noise. Recently, some GAN ~\cite{zhu2017unpaired,cc36} models are designed to learn the relationship between different views. Following this line, we consider to leverage GAN model for compensating the missing data. Nonetheless, directly applying GAN in solving PMVC is not straightforward. First, it is challenging for PMVC to generate the missing data based on partial views, rather than complete views. Second, it is under-explored for existing methods to explicitly consider the clustering task during learning representations from multiple views.


In this paper, we develop a novel generative partial multi-view clustering model, termed as GP-MVC, for the PMVC task. The proposed model is composed of four parts (See Fig. \ref{fig1}): multi-view encoder networks, weighted adaptive fusion layer, clustering layer, and view-specific generative adversarial networks. The proposed model employs multi-view encoder networks to encode the shared latent representation among multiple views. We naturally develop view-specific generative adversarial networks to predict the missing-view data conditioning on the latent representations from the other views. Specifically, we resort to adversarial training to explore consistent information among all the views. The generators of GP-MVC aim to complete the missing data, while the discriminators distinguish fake data from real ones for each view. One clustering layer is designed to boost the clustering structure of the common representation so that it could provide an explicit guidance for representation learning of clustering task. Moreover, we add a weighted adaptive fusion scheme to further exploit the complementary information among different views by introducing a group of learnable weights. By integrating clustering into the generating process, the proposed GP-MVC can adjust generator to compensate the ``ideal" missing data and thus improve the clustering performance.

This paper is an extension to our previous work \cite{partialWANG18}. Compared with \cite{partialWANG18}, several substantial differences have been made as follows: (1) We extend the architecture of GP-MVC from two views to multiple views to make our model generalize well in real-world applications. (2) We develop a new adaptive fusion layer for integrating the complementary information from different views. (3) More theoretical analyses, model discussions and experimental evaluations are provided.
We highlight the contribution of this work as the following.

\begin{itemize}
\item A novel GAN based partial multi-view clustering method named as GP-MVC is proposed to capture the shared clustering structure, and to generate missing-view data. Specifically, the proposed GP-MVC learns a consistent subspace shared by multiple views to provide common latent representations for both clustering and data generation tasks.
\item The proposed GP-MVC fully leverages consistent information provided by multi-view data. Particularly, GP-MVC obtains a latent representation with one view, with which it further generates the missing data of the corresponding views. The complementary missing-view data help improve clustering performance.
\item Extensive experiments on several multi-view datasets are conducted. Compared with several state-of-the-art methods, the experimental results proves the superiority of GP-MVC.
\end{itemize}

The remainder of this paper is organized as follows. In Section II, we conduct a brief review and analysis on related works. Then we introduce the proposed generative partial multi-view clustering in Section III. Experimental setting and evaluation results are reported in Section IV. Finally, we conclude our paper in Section V.

\section{Related Works}

As more and more missing multi-view data become common in real-world application. Incomplete multi-view clustering methods have been proposed for multi-view data clustering. In this section, we will introduce some multi-view clustering methods, partial multi-view clustering methods, and generative adversarial networks.

\vspace{-0.3cm}
\subsection{Multi-View Clustering}
Multi-view clustering methods can be divided into three categories. The first category is spectral-based methods~\cite{cc1,cc14,cc16,huang2019multi}. These methods usually learn a shared similarity matrix among different views and conduct spectral clustering for the final partition result. For example, Kumar \emph{et al.}~\cite{cc14} designed a co-regularized multi-view spectral clustering, which can perform clustering on different views simultaneously. Motivated by this work, Tsivtsivadze \emph{et al.}~\cite{cc16} designed neighborhood co-regularized multi-view spectral clustering for microbiome data clustering. The second one is subspace-based method which learns a shared coefficient matrix from each view~\cite{cc18,wang2017exclusivity,luo2018consistent}. Based on this idea, Yin \emph{et al.}~\cite{cc18} proposed a pairwise sparse multi-view subspace clustering by enforcing the coefficient matrices from each pair of views as similar as possible. Different from the above approaches, the third category~\cite{cc10} mainly uses non-negative matrix factorization to learn a common indicator matrix from different views. Zhao \emph{et al.}~\cite{cc10} adopted a deep semi-nonnegative matrix factorization to perform multi-view clustering. On account of increasing application of partial multi-view clustering, researchers have proposed partial multi-view clustering.

\vspace{-0.3cm}
\subsection{Partial Multi-View Clustering}
Piyush \emph{et al.}~\cite{cc31} designed the first partial multi-view clustering approach. They adopted one views kernel representation as the similarity matrix, and employed Laplacian regularization to complete the kernel matrices of incomplete view. Nonetheless, this approach requires one complete view that consists of all instances. To tackle the problem, an incomplete multi-view clustering was developed based on kernel canonical correlation analysis~\cite{cc29,liu2018late}. These methods optimize the alignment of shared instances in the dataset and thereby can collectively complete the kernel matrices of incomplete view. Despite the effectiveness of these methods, they can only be applied in kernel-based methods. Recently, Non-negative Matrix Factorization (NMF) based Partial View Clustering (PVC) algorithm was proposed in~\cite{cc25}. It establishes a latent subspace in which different view's examples belonging to one instance are close to each other. It shows to be effective for partial multi-view data. Inspired by its promising performance, numerous NMF based multi-view methods are developed~\cite{cc26,cc27}. For example, Rai \emph{et al.}~\cite{cc27} improve it with graph regularized NMF. Zhao \emph{et al.}~\cite{cc26} proposed an Incomplete Multi-Modal Visual Data Grouping (IMG), and its main idea is to learn a unified framework by integrating latent subspace generation and compact global structure. However, all these methods utilize a latent space learned for multi-view data with NMF, which restricts its application over negative feature data and nonnegative matrix factorization requires complicated calculation, so they cannot be used for large-scale data. On the other hand, a lot of works using deep model to generate images and achieve successful paradigms.

\vspace{-0.3cm}
\subsection{Generative Adversarial Networks}
Generative adversarial network (GAN) was developed by Goodfellow \emph{et al.}~\cite{cc37}. It quickly attracted a huge amount of interest because of its extraordinary performance and interesting theory. Recently, many variations of GAN have been put forward for various goals and applications. For example, Mao \emph{et al.} proposed Least Squares GAN to solve the vanishing gradients problem when minimizing the objective function~\cite{mao2017least}. Rather than cross entropy loss function, it adopts least squares loss function for the discriminator. Zhang \emph{et al.} \cite{zhang2017stackgan} applied GAN in photorealistic images generation conditioned on text descriptions, and introduced Stacked GAN (StackGAN). In Stack-GAN, two GANs in different stages are adopted to generate high-resolution images. Odena \emph{et al.} introduced more structures and a specialized cost function to GAN latent space for high-quality sample generation, and then proposed conditional GAN associated with an auxiliary classifier (AC-GAN)~\cite{odena2016conditional}. In this way, the conditional information can be class labels or data from other modalities. Isola \emph{et al.} \cite{isola2017image} further explored the application of conditional GANs on paired training data and developed pix2pix GAN, and it can transfer images from one distribution to another effectively. Zhu \emph{et al.} proposed Cycle GAN~\cite{zhu2017unpaired}, which trains unpaired image with a cycle consistent adversarial network by adding cycle consistent loss. It shows a more powerful ability than pix2pix GAN  in image translation from one domain to another, and hence effectively solves the paired sample shortage problem.  GAN also gains a wide application in multi-view data generation~\cite{cc36} and~\cite{zhao2017multi}.

\vspace{-0.2cm}
\section{Generative Partial Multi-View Clustering}
\subsection{Motivation}
Existing works which can deal with partial multi-view data are mostly based on kernel and NMF to predict the missing data and then do clustering task at the same time. Despite appealing performance achieved, they still have two limitations. First, non-negative matrix factorization based methods need a lot of computation for inverse operation. As a result, they cannot be applied to large-scale data. Second, all these methods only focus on learning a shared latent space for clustering. They ignore learning a latent space that is suitable for clustering and can be used to generate the missing view data simultaneously. To address these two challenges, we design a novel model called Generative Partial Multi-View Clustering (GP-MVC). We combine the  generative capacity of GAN and the clustering capacity of deep embedding clustering to our model. Thus, it can generate the missing-view data and learn a better clustering structure for partial multi-view data at the same time.

\begin{figure}[!t]
\setlength{\abovecaptionskip}{0.2cm}
  \setlength{\belowcaptionskip}{0.1cm}
  \centering
  \includegraphics[width=3.5in]{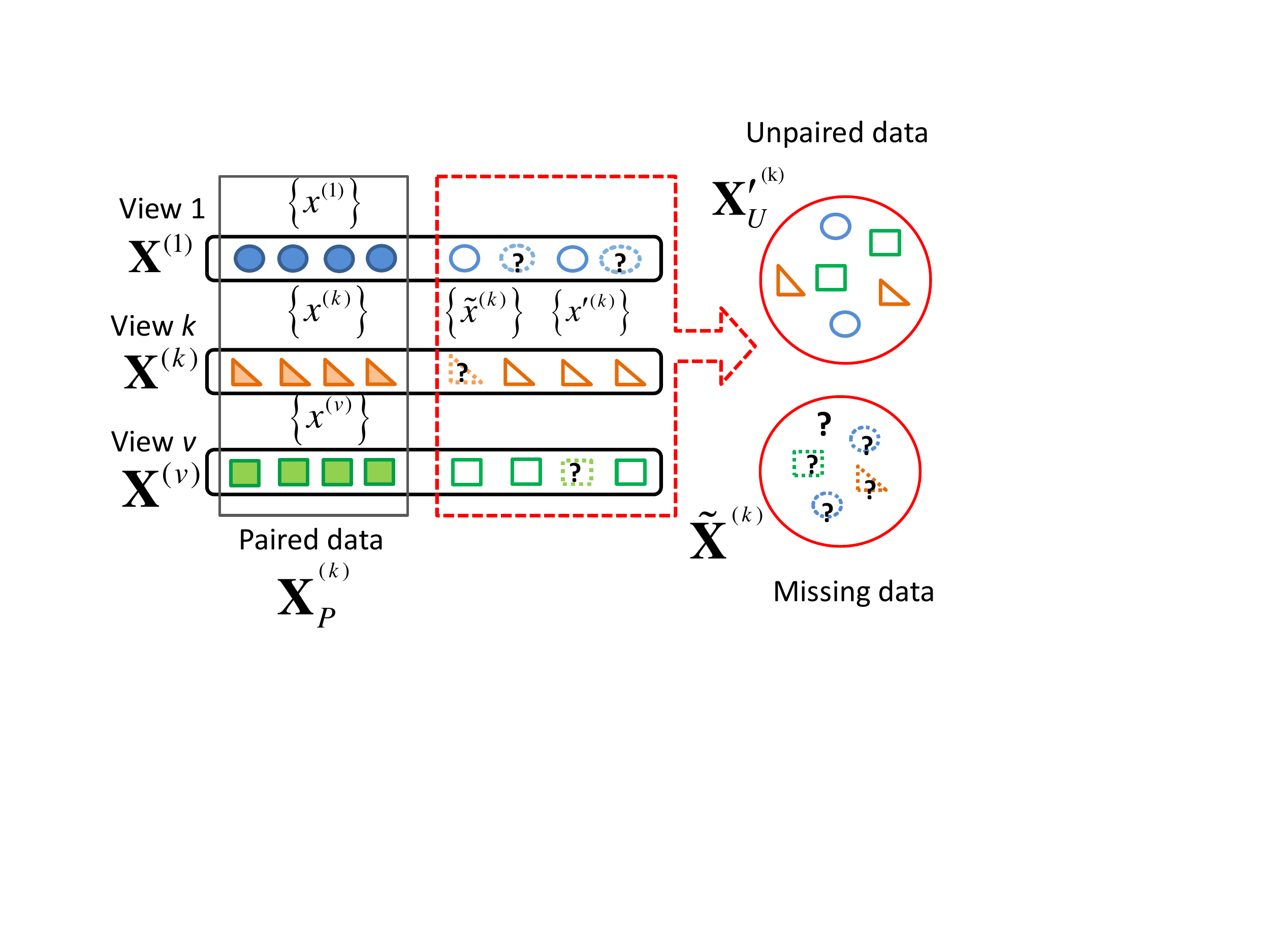}
  \vspace{-0.7cm}
  \caption{Illustration for partial multi-view data, the data in solid box is complete paired data, while them in red dash box is partial data.}\label{fig2}
  \vspace{-0.2cm}
\end{figure}

\textbf{Notations.}
We represent the data with the multi-view data matrix ${\bf{X}} = \left\{ {{{\bf{X}}^{(1)}},\;{{\bf{X}}^{(2)}},\cdots,{{\bf{X}}^{(V)}}}\right\}$, where ${{\bf{X}}^{(v)}}=\left\{ {{x_1^{(v)}},\;{x_2^{(v)}},\cdots,{x_N^{(v)}}} \right\} \in {{\bf{R}}^{N \times {d_v}}}(v = 1,2,\cdots,V)$, $V$ is the number of views, ${\bf{N}}$ is the number of samples, and ${d_v}$ is the feature dimension of $v$-th view. Since the setting of our model is partial multi-view clustering, we divide the multi-view data ${\bf{X}}$ to two parts: one is paired data $\left\{{{x^{(1)}},{x^{(2)}},\cdots,{x^{(V)}}} \right\}$ in which all the view is complete. The other one is unpaired data $\left\{ {{{x'}}^{(1)}}, {{{x'}}^{(2)}},\cdots, {{{x'}}^{(V)}}\right\}$ in which some data is missing. We give an example of multi-view data in Fig. \ref{fig2}. $\left\{ {{{\tilde x}}^{(1)}}, {{{\tilde x}}^{(2)}},\cdots, {{{\tilde x}}^{(V)}}\right\}$ respectively denote the missing data or generated data of each view.

\vspace{-0.3cm}
\subsection{Framework}
\textbf{Network Architecture.}
Fig. \ref{fig1} illustrates the architecture of our model for partial multi-view data. It is composed of five sub-networks: encoder network ${{\bf{E}}}$, weighted adaptive fusion layer, deep embedding clustering layer, generator network ${{\bf{G}}}$, and discriminator network ${{\bf{D}}}$. For multi-view data, corresponding to each view, our model has $V$ encoders, one fusion layer, one clustering layer, $V$ generators and $V$ discriminators. We introduce the model in details as follows.

\emph{Encoder network} ${{\bf{E}}}$: ${{\bf{R}}^{{d_v}}} \to {{\bf{R}}^m}$. Each view has a encoder which is stacked fully connected. It encodes the $v$-th original view ${{\bf{X}}^{(v)}}$ to a latent representation ${\bf{Z}}^{(v)} \in {{\bf{R}}^{N \times m}}(v=1,2,\cdots,V)$, where ${{\bf{Z}}^{(v)}}{\rm{ = }}\left\{ {z_1^{(v)},z_2^{(v)}, \cdots, z_N^{(v)}} \right\}$. Denote the nonlinear function of $v$-th encoder by ${{\bf{E}}_v}$. It maps the ${d_v}$-dimensional original data ${x_i^{(v)}}$ to a $m$-dimensional latent representation ${z_i^{(v)}}$, ${z_i^{(v)}} = {{\bf{E}}_v}({x_i^{(v)}};\theta )$, where $\theta$ is shared parameters of all encoders. In order to capture shared structure of multi-view data, we partially share parameters of encoders for all the view, i.e. $\theta$.

\emph{Generator network} ${{\bf{G}}}$: ${{\bf{R}}^m} \to {{\bf{R}}^{{d_v}}}$. In our model, the generator can also be seen as decoder as it has symmetrical structure with encoder, i.e., stacked fully connected. Corresponding to each view, there is one exclusive generator which can generate the corresponding view. For example, the $v$-th generator ${{\bf{G}}_v}$ inputs the latent representation ${{\bf{Z}}^{(w)}}\;(w=1,2,\cdots,V,w\neq v)$ and outputs the generated $v$-th view ${{\bf{\tilde X}}^{(v)}}$. That is to say, for ${{\bf{G}}_v}$, it will outputs ${{{\tilde x}}_i^{(v)}} = {{\bf{G}}_v}({{{z}}_i^{(w)}} )$ , no matter which latent representation ${{\bf{Z}}_w}$ it inputs. Thus, we hope the latent representation ${{\bf{Z}}^{(1)}},{{\bf{Z}}^{(2)}},\cdots,{{\bf{Z}}^{(V)}}$ are similar to each other. The best status is they are equal with each other. However, the equal condition is too strict, so we introduce a common representation ${{\bf{Z}}}$ for each view to relax the equal condition.

\emph{Discriminator network} ${{\bf{D}}}$: ${{\bf{R}}^{{d_v}}} \to \{ 0,1\} $. Similar with generator, corresponding to each view, there is one exclusive discriminator which is composed of 3 stacked fully connected layers. As shown in Fig. \ref{fig1}, all discriminators are connected with generators. Take the $v$-th discriminator ${{\bf{D}}_v}$ as an example, it inputs the real sample ${x_i^{(v)}}$ and the generated fake sample ${{\tilde x}_i^{(v)}}$ of the $v$-th view. Then it outputs judgment result of the authenticity for the generated sample ${{\tilde x}_i^{(v)}}$: real/fake, i.e., $0/1 = {{\bf{D}}_v}({{{\tilde x}}_i^{(v)}},{{{ x}}_i^{(v)}} )$. The result means that discriminator ${{\bf{D}}_v}$ considers the generated sample ${{{\tilde x}}_i^{(v)}}$ is real or fake. The discriminator's result will also feedback to generator ${{\bf{G}}_v}$, and prompt generator to produce more realistic sample. The process will be repeat until generator can produce so real sample ${{\tilde x}_i^{(v)}}$ that discriminator cannot distinguish which one is generated sample.

\emph{Weighted adaptive fusion layer}: After encoder, we got $V$ latent spaces ${{\bf{Z}}^{(1)}},{{\bf{Z}}^{(2)}},\cdots,{{\bf{Z}}^{(V)}}$. To fully explore the complimentary information across multi-view images, we adaptively fuse the latent representations of different view and learn a common representation ${\bf{Z}}$. Specifically, we design a learnable fusion layer to obtain ${\bf{Z}}$ by ${\bf{Z}} = f(\{ {{\bf{Z}}^{(v)}}\} _{v = 1}^V;\beta )$, where $f(\bullet \ ; \beta)$ denotes the fusion function parameterized by $\beta=[\beta_1,\dots,\beta_V]$.

\emph{Deep embedded clustering layer}: This layer will improves the distribution of common representation ${{\bf{Z}}}$ and obtain clustering result. First, we compute the original distribution of common representation ${{\bf{Z}}}$, named it as ${{\bf{P}}}$, then based on ${{\bf{P}}}$, we can compute a target distribution ${{\bf{Q}}}$ which is more compact and suitable for clustering. According target distribution, we refine encoder network ${{\bf{E}}}$ and hope it can learn a latent representation which is similar to the target distribution.

\vspace{-0.2cm}
\subsection{Objective Function}
Our objective function includes four terms as: auto-encoder loss, adversarial training loss, weighted adaptive fusion loss and KL clustering loss.

\subsubsection{Auto-Encoder Loss}

The auto-encoder loss works on encoder network ${{\bf{E}}}$ and generator network ${{\bf{G}}}$. We hope the output of generator is similar with the input of encoder. Thus, we minimizes the squared F-norm of the reconstruction error between the generated sample and input sample. The auto-encoder loss is
\vspace{-0.3cm}
\begin{equation}\label{eq2}
{{L_{{\rm{AE}}}} = \mathop {\min }\limits_{\begin{array}{*{20}{c}}
{\theta ,{{\bf{E}}_1}, \cdots ,{{\bf{E}}_V},}\\
{\;\;\;{{\bf{G}}_1}, \cdots ,{{\bf{G}}_V}}
\end{array}} \sum\limits_{v = 1}^V {\left\| {{{\bf{X}}^{(v)}}{\rm{ - }}{{\bf{G}}_v}({{\bf{E}}_v}({{\bf{X}}^{(v)}};\theta ))} \right\|_{\rm{F}}^2} }
\end{equation}


Take the $v$-th view ${{\bf{X}}^{(v)}}$ as an example, when it passes through encoder ${{\bf{E}}_v}$,  we get ${{\bf{E}}_v}({{\bf{X}}^{(v)}};\theta )$, i.e., the latent representation ${{\bf{Z}}^{(v)}}$.  In auto-encoder loss, we take generator ${{\bf{G}}^{(v)}}$ as decoder corresponding to encoder ${{\bf{E}}^{(v)}}$. Thus, the function of generator is reconstructing input sample from latent representation ${{\bf{Z}}^{(v)}}$. ${{\bf{G}}_v}({{\bf{E}}_v}({{\bf{X}}^{(v)}};\theta ))$ denotes the output of generator. We minimize the reconstruction error for obtain encoder and generator networks which can output similar sample with the input sample.

Generally, when all input data are paired, auto-encoder loss alone is sufficient to train the encoder and generator networks. Nonetheless, in partial multi-view learning, the performance of encoder and generator networks will be greatly degraded because of unpaired data. To tackle this problem due to unpaired data, apart from auto-encoder loss, we further employ adversarial training loss to refine the network in the objective function.
\begin{figure}[!t]
\setlength{\abovecaptionskip}{0.2cm}
  \setlength{\belowcaptionskip}{0.1cm}
  \centering
  \includegraphics[width=8cm]{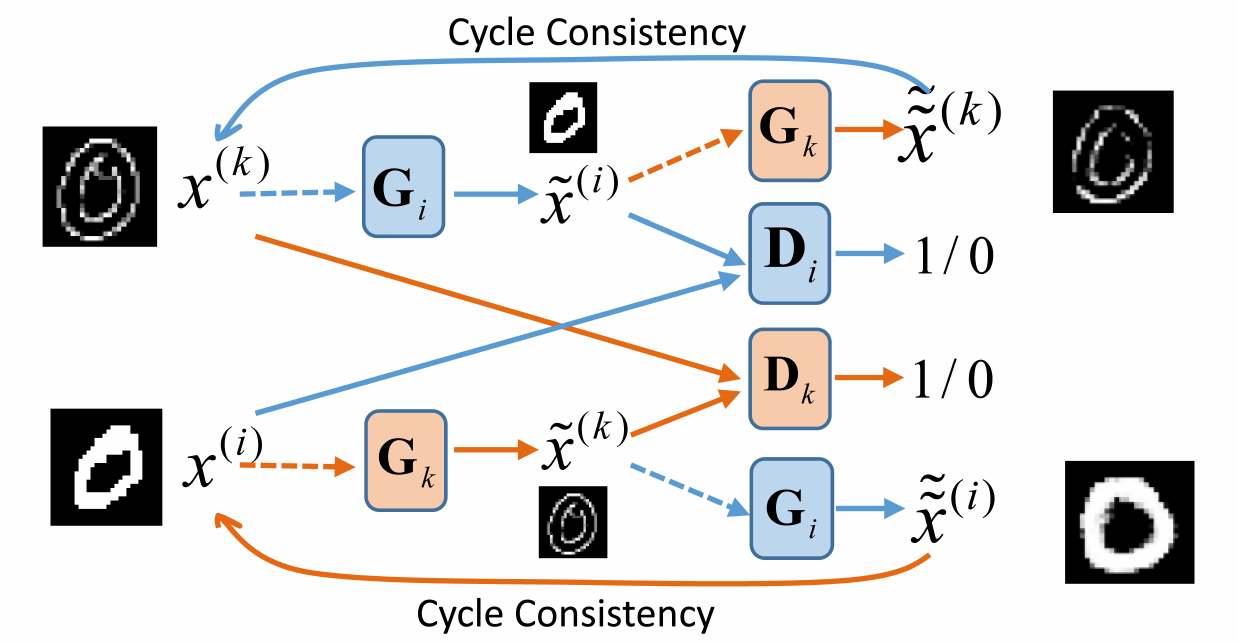}
   \vspace{-0.2cm}
  \caption{The framework of cycle GAN. We take view $i$ and view $k$ as an example.}\label{fig21}
  \vspace{-0.3cm}
\end{figure}

\subsubsection{Adversarial Training Loss}
Suppose $x$ is a sample from data distribution $P_{\rm data}$, and $z$ is a noise sample from noise distribution $P_{\rm z}$. A typical GAN is composed of two sub-networks, namely, a generator ${\bf{G}}$ and a discriminator ${\bf{D}}$. Among them, ${\bf{G}}$ generates a fake image ${\bf{G}}(z)$ with a vector of random noise $z$ as input, while ${\bf{D}}$ aims to distinguish the fake image generated by ${\bf{G}}$ from the real image, and it will return a value ranging from $0$ to $1$, which indicates the probability whether the input image is real or fake. GAN adopts the idea of game theory, and ${\bf{G}}$ and ${\bf{D}}$ are somehow participating a min-max game. Specifically, the loss functions of ${\bf{G}}$ and ${\bf{D}}$ respectively try to minimize and maximize the likelihood that the fake image assigns to the fake source. From this view, we can easily understand the loss function of GAN
\begin{equation}\label{eq01}
\begin{array}{l}
{L_{\rm GAN}}({{\bf{G}}},{{\bf{D}}}) = \mathop {\min }\limits_{\bf{G}} \mathop {\max }\limits_{\bf{D}} {\mathbb{E}_{x\sim{P_{data}}}}\left[ {\log {\bf{D}}(x)} \right]\\
\;\;\;\;\;\;\;\;\;\;\; + {\mathbb{E}_{z\sim{P_z}}}\left[ {\log (1 - {\bf{D}}({\bf{G}}(z)))} \right].
\end{array}
\end{equation}

Considering that there exist a large amount of unpaired data and lack paired data in our setting, we employ cycle GAN in our model to conduct adversarial learning, which can effectively handle the problem due to insufficiency of paired data. A cycle GAM model trains two GAN models and adds a GAN loss and a cycle consistency loss based on Eq. (\ref{eq01}) to tackle unpaired data situation. Fig. \ref{fig21} shows the framework of a cycle GAN model. Its main idea is each data distribution can generate the other via these two GAN models. According to the theory of cycle GAN, we design a multi-view adversarial training loss for multi-view data in our model
\begin{equation}\label{eq31}
\begin{array}{l}
{L_{{\rm{AT}}}}({{\bf{G}}_1},{{\bf{D}}_1}, \cdots ,{{\bf{G}}_V},{{\bf{D}}_V}) = \sum\limits_{v = 1}^V {\mathop {\min }\limits_{{{\bf{G}}_v}} \mathop {\max }\limits_{{{\bf{D}}_v}} {L_{{\rm{GAN}}}}({{\bf{G}}_v},{{\bf{D}}_v})} \\
\;\;\;\;\;\;\;\;\;\;\;\;\;\;\;\;\;\;\;\;\;\;\;\;\;\;\;\;\;\;\;\;\;\;\;\;\;\;\;\;\;\;\;\;\;\;\;{\rm{ + }}{L_{{\rm{cyc}}}}({{\bf{G}}_1}, \cdots {{\bf{G}}_V}).
\end{array}
\end{equation}


The adversarial training loss mainly works on generator and discriminator networks. Next we will give the definition of GAN loss ${L_{\rm GAN}}$ and cycle consistency loss ${L_{\rm cyc}}$. Suppose that the data distribution of the $v$-th views is ${x^{(v)}} \sim P({{\bf{X}}^{(v)}})$, and let ${{\bf{G}}_v} \circ {{\bf{E}}_w}{\rm{ = }}{{\bf{G}}_v}({{\bf{E}}_w}({x^{(w)}},\theta ))$ denote the mapping from $w$-th view data distribution $P({{\bf{X}}^{(w)}})$ to $v$-th view data distribution $P({{\bf{X}}^{(v)}})$, i.e., using $w$-th view sample ${x^{(w)}}$ to generate the $v$-th view sample ${{\tilde x^{(v)}}}\sim P({{\bf{X}}^{(v)}})$ while ${x^{(v)}}$ is paired with ${x^{(w)}}$. The same theory for ${{\bf{G}}_w} \circ {{\bf{E}}_v}{\rm{ = }}{{\bf{G}}_w}({{\bf{E}}_v}({x^{(v)}},\theta ))$, which transform the $v$-th view sample to $w$-th view sample. ${{\bf{D}}_v}$ inputs ${{\bf{G}}_v}$'s generated sample ${{\tilde x^{(v)}}}$ and real sample ${x^{(v)}}$, and then outputs the discriminant result. 
Thus, for partial multi-view data, the loss of $v$-th GAN network in our model is
\begin{equation}\label{eq3}
\begin{array}{l}
{L_{\rm GAN}}({{\bf{G}}_v},{{\bf{D}}_v})\!=\! \mathop {\min }\limits_{{{\bf{G}}_v}} \mathop {\max }\limits_{{{\bf{D}}_v}}{\mathbb{E}_{{{x^{(v)}}}\sim P({{\bf{X}}^{(v)}})}}\left[ {\log \;{{\bf{D}}_v}({{x^{(v)}}})} \right]\\
\;\;\;\;\; + {\sum\limits_{w = 1,w \neq v}^V {{\mathbb{E}_{{{x^{(w)}}}\sim P({{\bf{X}}^{(w)}})}}\left[ {\log \;(1 - {{\bf{D}}_v}({{\bf{G}}_v}\circ{{{\bf{E}}_w}}({{x^{(w)}}})))} \right]}}.\\
\end{array}
\end{equation}

The generator aims at generating fake sample which is similar to real sample, and the discriminator tries to distinguish the generated sample from real sample. In this way, generator and discriminator play an opposite game until convergence when generator can generate great real sample. Nonetheless, GAN has a character that it maps a same input to any random permutation of sample in the target data distribution. Therefore, the model cannot obtain a desired output only with the GAN loss. To tackle this problem, cycle GAN further employs cycle-consistent loss to update the learned mapping and thereby reduces the space of possible mapping functions. Specifically, for each sample ${x^{(v)}}$ in $v$-th views, after passing through encoder ${{\bf{E}}_v}$ and generator ${{\bf{G}}_w}$, we get the generated sample ${{\tilde x^{(w)}}}$. Then, let the generated sample ${{\tilde x^{(w)}}}$ pass through encoder ${{\bf{E}}_w}$ and generator ${{\bf{G}}_v}$. Finally, we obtain the generated sample ${{{\tilde{\tilde x}}^{(v)}}}$ after passing the translation cycle. Since the cycle consistency loss assists the generator in mapping a given sample ${x^{(v)}}$ to a desired output ${\tilde x^{(w)}}$ which is special and paired with ${x^{(v)}}$, we can combine GAN loss and cycle consistency loss to guarantee a desired output effectively. The multi-view cycle consistency loss of our model is minimizing the $\ell_1$-norm of the reconstruction error between the final generated sample and input sample
 \vspace{-0.3cm}
\begin{equation}\label{eq33}
\begin{array}{l}
{L_{\rm cyc}}({{\bf{G}}_1}, \cdots {{\bf{G}}_V}) = \mathop {\min }\limits_{{{\bf{G}}_1}, \cdots {{\bf{G}}_V}}\{\sum\limits_{v = 1}^V {\sum\limits_{w = 1,w \neq v}^V }\\
\;\;\;\;\;\;{{\mathbb{E}_{{x^{(v)}}\sim P({{\bf{X}}^{(v)}})}}\left\| {{{\bf{G}}_v} \circ {{\bf{E}}_w}({{\bf{G}}_w} \circ {{{\bf{E}}_v}}({x^{(v)}}))-{x^{(v)}}} \right\|_1}\}.
\end{array}
\end{equation}
\vspace{-3mm}


\subsubsection{Weighted Adaptive Fusion Loss}

Before clustering, we need fuse all the latent space. There we use weighted adaptive fusion
We will learn $v$ latent subspaces for each view data $\left\{ {{\bf{Z}}^{(1)}},{{\bf{Z}}^{(2)}},\cdots,{{\bf{Z}}^{(V)}} \right\}$, where ${\bf{Z}}^{(v)} = {{\bf{E}}_v}({{\bf{X}}^{(v)}};\theta )$. Then, by the following equation, we get a common representation ${{\bf{Z}}}$ based on all latent subspaces
\begin{equation}\label{eq4}
{\bf{Z}} = h({{\bf{Z}}^{(1)}},\cdots,{{\bf{Z}}^{(V)}}),
\end{equation}
where $h(\cdot)$ denotes a concatation or summation function.

After getting the latent representation ${\bf{Z}}^{(v)}$ of $V$ views from encoder, we adopt an adaptive fusion method to extract public identification information that is beneficial to clustering from the multi-modal primitive space. In a manner similar to classification, the authentication information is used to approximate the ideal data distribution. We define the adaptive fusion loss function as
\begin{equation}\label{eq111}
\begin{array}{l}
{L_{FU}} = \mathop {\min }\limits_\beta  \left\| {{\bf{Z}} - \sum\limits_{v = 1}^V {{\beta _v}{{\bf{Z}}^{(v)}}} } \right\|_{\rm{F}}^2\\
{\kern 1pt} {\kern 1pt} {\kern 1pt} {\kern 1pt} {\kern 1pt} {\kern 1pt} {\kern 1pt} {\kern 1pt} {\kern 1pt} {\kern 1pt} {\kern 1pt} {\kern 1pt} {\kern 1pt} {\kern 1pt} {\kern 1pt} {\kern 1pt} {\kern 1pt} {\kern 1pt} {\kern 1pt}  = \mathop {\min }\limits_\beta  \left\| {f(\{ {{\bf{Z}}^{(v)}}\} _{v = 1}^V;\beta ) - \sum\limits_{v = 1}^V {{\beta _v}{{\bf{Z}}^{(v)}}} } \right\|_{\rm{F}}^2
\end{array}
\end{equation}
where ${{\bf{Z}}^{(v)}}={{\bf{E}}_v}({{\bf{X}}^{(v)}},{\theta})$ is the output of encoder, $\beta  = \{ {\beta _1}, \cdots ,{\beta _V}\}$ is a group learnable parameter, and $f(\bullet \ ; \beta)$ denotes the fusion function.

\subsubsection{KL Clustering Loss}
The analysis demonstrates that the performance of the generator and discriminator networks can be improved by employing adversarial training loss. Nevertheless, it makes little modification to the encoders. The encoders learn a common representation ${\bf{Z}}$ for the final clustering task based on paired data and unpaired data. However, unpaired data will negatively affect its clustering performance. To optimize encoder and obtain a better clustering structure, we add a clustering loss measured by Kullback-Leibler divergence (KL-divergence) in our model. We represent $k$ initial clustering centroids with $\left\{ {{\mu _j}} \right\}_{j = 1}^k$. In order to measure the similarity between common representation point ${z_i}$ and centroid ${\mu _j}$, we refer to the method in \cite{xie2016unsupervised} and employ the Student's $t$-distribution as a kernel. Then we can calculate the probability that a sample $i$ is assigned to cluster $j$ with the following formula
 \vspace{-0.3cm}
\begin{equation}\label{eq5}
 {q_{ij}} = \frac{{{{(1 + {{\left\| {{z_i} - {\mu _j}} \right\|}^2}/\alpha )}^{ - \frac{{\alpha  + 1}}{2}}}}}{{\sum\nolimits_{j'} {{{(1 + {{\left\| {{z_i} - {\mu _{j'}}} \right\|}^2}/\alpha )}^{ - \frac{{\alpha  + 1}}{2}}}} }}.
 \vspace{-1mm}
\end{equation}

It is also called soft assignment. Herein, we denote the degree of freedom of the Student's $t$-distribution as $\alpha$. Then we first raise ${q_i}$ to the squared and then normalize it with frequency per cluster to obtain the target distribution ${p_i}$.
\begin{equation}\label{eq6}
{p_{ij}} = \frac{{{{q_{ij}^2} \mathord{\left/
 {\vphantom {{q_{ij}^2} {{f_j}}}} \right.
 \kern-\nulldelimiterspace} {{f_j}}}}}{{\sum\nolimits_{j'} {{{q_{ij'}^2} \mathord{\left/
 {\vphantom {{q_{ij'}^2} {{f_{j'}}}}} \right.
 \kern-\nulldelimiterspace} {{f_{j'}}}}} }},
 \vspace{-1mm}
\end{equation}
where ${f_j} = \sum\nolimits_i {{q_{ij}}}$ represents soft cluster frequency. In this way, our method improves clustering performance and is able to lay special stress on data points assigned with high confidence. Finally, we leverage minimizing the KL-divergence between original data distribution and target distribution as clustering loss
\begin{equation}\label{eq7}
\vspace{-1mm}
{L_{KL}}\;{\rm{ =  KL(P||Q)  =  }}\sum\limits_i {\sum\limits_j {{p_{ij}}\log \frac{{{p_{ij}}}}{{{q_{ij}}}}} }.
\vspace{-1mm}
\end{equation}

We aim to match the soft assignment ${q_i}$ to the target distribution ${p_i}$. This assists in sharpening the data distribution and also concentrating the same class data. Additionally, we are able to achieve a more common representation that works more effectively in partial multi-view clustering. 

\subsubsection{Overall objective}

By integrating auto-encoder loss, adversarial training loss, weighted adaptive fusion loss and KL clustering loss, we have the following objective function of GP-MVC
\begin{equation}\label{eq1}
L = \mathop {{\rm{min}}}\limits_{\scriptstyle\theta ,{{\bf{E}}_1}, \cdots ,{{\bf{E}}_V},\hfill\atop
\scriptstyle\;\;{{\bf{G}}_1}, \cdots ,{{\bf{G}}_V}\hfill} \;\mathop {{\rm{max}}}\limits_{{{\bf{D}}_1}, \cdots ,{{\bf{D}}_V}} \;{L_{{\rm{AE}}}} + {\lambda _1}{L_{{\rm{AT}}}}  + {\lambda _2}{L_{{\rm{FU}}}}+ {\lambda _3}{L_{{\rm{KL}}}},
\end{equation}
where ${\lambda _1}$, ${\lambda _2}$, and ${\lambda _3}$ are parameters to adjust the impact of each term in all objective function. Next we will give the details of each term in Eq. (\ref{eq1}).

\begin{algorithm}[!t]
  \caption{Generative Partial Multi-view Clustering}
  \label{alg::conjugateGradient}
  \begin{algorithmic}[1]
    \Require
      A multi-view data matrix ${\bf{X}} = \left\{ {{{\bf{X}}^{(1)}},\;{{\bf{X}}^{(2)}},\cdots,{{\bf{X}}^{(V)}}}\right\}$; Parameters ${\lambda_1}$, ${\lambda_2}$, ${\lambda_3}$;
    \Ensure
      The result of clustering.
    \State \textbf{Initialize:} The parameter for all models: encoder ${\bf{E}}$, clustering layer, generator ${\bf{G}}$, and discriminator ${\bf{D}}$.
    \State \textbf{Step 1:} Train encoder ${\bf{E}}$ and generator ${\bf{G}}$;
    \For{each $ite\in$ pre-specified iterations}
      \State Input paired data $\left\{{{x^{(1)}},{x^{(2)}},\cdots,{x^{(V)}}} \right\}$;
      \State Update ${\bf{E}}$ and ${\bf{G}}$ by Eq. \eqref{eq2};
      \State Computer the clustering centroids $\left\{ {{\mu _j}} \right\}_{j = 1}^k$;
    \EndFor
    \State \textbf{Step 2:} Train generator ${\bf{G}}$ and discriminator ${\bf{D}}$;
    \For{each $ite\in$ pre-specified iterations}
      \State Input all data ${\bf{X}}$;
      \State Update ${\bf{G}}$, and ${\bf{D}}$ by Eq. \eqref{eq31};
      \State Generate missing sample $\left\{{{\tilde x^{(1)}},{\tilde x^{(2)}},\cdots,{\tilde x^{(V)}}} \right\}$;
      \State Computer the common representation ${{\bf{Z}}}$;
    \EndFor
    \State \textbf{Step 3:} Train all model;
    \For{each $ite\in$ pre-specified iterations}
      \State Input the clustering centroids $\left\{ {{\mu _j}} \right\}_{j = 1}^k$, the common representation ${{\bf{Z}}}$, and the completed data;
      \State Update ${\bf{E}}$, ${\bf{G}}$, and ${\bf{D}}$ by Eq. \eqref{eq1};
      \State Computer the common representation ${{\bf{Z}}}$;
    \EndFor
    \State Clustering on the common representation ${{\bf{Z}}}$;
  \end{algorithmic}
\end{algorithm}
\vspace{-0.3cm}

\subsection{Implementation}

Step 1: Training encoder $\left\{ {{\bf{E}}_1}, \cdots, {{\bf{E}}_V} \right\}$ and generator $\left\{ {{\bf{G}}_1}, \cdots,{{\bf{G}}_V} \right\}$ on paired data.

We only use paired data to train encoder and generator of our model at first. Since they can be seen as auto-encoder network, we just use the auto-encoder loss to optimise $\left\{ {{\bf{E}}_1}, \cdots,{{\bf{E}}_V}, {{\bf{G}}_1},\cdots,{{\bf{G}}_V}\right\}$. Specifically, we take paired data $\left\{{{x_i^{(1)}},{x_i^{(2)}},\cdots,{x_i^{(V)}}} \right\}$ as input for encoder ${{\bf{E}}_1},\cdots, {{\bf{E}}_V}$ and get $V$ latent subspace $\left\{ {{\bf{Z}}^{(1)}},{{\bf{Z}}^{(2)}},\cdots,{{\bf{Z}}^{(V)}} \right\}$. Then they pass through generator $ {{\bf{G}}_1}, \cdots,{{\bf{G}}_V} $ respectively and output $\left\{{{\tilde x_i^{(1)}},{\tilde x_i^{(2)}},\cdots,{\tilde x_i^{(V)}}} \right\}$.  Finally we compute clustering centroids $\left\{ {{\mu _j}} \right\}_{j = 1}^k$ and use auto-encoder loss to update encoder and generator networks. By step 1, we obtain clustering centroids which will be used in step 3. These clustering centroids learned by paired data can help the generated sample for missing-view sample to be assigned to the right group.

Step 2: Training generator $\left\{ {{\bf{G}}_1}, \cdots,{{\bf{G}}_V} \right\}$ and discriminator $\left\{ {{\bf{D}}_1}, \cdots,{{\bf{D}}_V} \right\}$ on all data.

We use all data to train generator and discriminator networks, i.e., multi-view cycle GAN, in this step. For paired data $\left\{{{x_i^{(1)}},\;{x_i^{(2)}},\;\cdots,\;{x_i^{(V)}}} \right\}$, we directly take them as input of encoder $\left\{ {{\bf{E}}_1}, \cdots,{{\bf{E}}_V}\right\}$. When encountering unpaired data $\left\{ {{{x'}_j}^{(v)}}\right\}$ of $v$-th view, we randomly choose one sample from $w$-th view ($w\neq v$) $\left\{ {{{x}_i}^{(w)}},{{{x'}_j}^{(w)}}\right\}$ as input of encoder in each epoch to increase the number of unpaired data. After step 2, we save the output of generator $\left\{{{\tilde x_j^{(1)}},{\tilde x_j^{(2)}},\cdots,{\tilde x_j^{(V)}}} \right\}$, i.e., the generated sample for missing view. Then we compute a new common representation ${{\bf{Z}}}$ on complete database.

Step 3: Training all model on the completed dataset.


We use the clustering centroids $\left\{ {{\mu _j}} \right\}_{j = 1}^k$ from step 1, and the completed dataset $\left\{ {\left\{ {x_i^{(1)}, \cdots ,x_i^{(V)}} \right\},} \right.$ $\left. {\left\{ {{{x'}_j}^{(1)}, \cdots ,\tilde x_j^{(V)}} \right\}, \cdots ,\left\{ {\tilde x_j^{(1)}, \cdots ,{{x'}_j}^{(V)}} \right\}} \right\}$ from step 2 as input for our model to train it again. In each epoch, we update the the clustering centroids, the common representation and the generated sample for the missing view.

\emph{Algorithm} 1 illustrates the training procedure of our model.


\begin{table*}[t]
\begin{center}
\caption{The average Clustering Accuracy in terms of different impartial ratios on the BDGP Database.}
\scalebox{1}{
\begin{tabular}{ccccccc}
  \toprule
{\textbf{Methods}}&\textbf{0.1}&{\textbf{0.3}}&\textbf{0.5}& {\textbf{0.7}}& {\textbf{0.9}}\\
\midrule
best SC&0.4748$\pm$0.0131&0.5169 $\pm$0.0174&0.5692$\pm$0.0159&0.6139$\pm$0.0121&0.6716$\pm$0.0136\\
AMGL~\cite{ref13}&0.2524$\pm$0.0349&0.2357$\pm$0.0180 &0.2538$\pm$0.0155&0.2807$\pm$0.0125&0.2958$\pm$0.0195\\
RMSC~\cite{cc17}&0.3395$\pm$0.0050&0.3683$\pm$0.0051&0.3907$\pm$0.0045&0.4233$\pm$0.0048&0.4499$\pm$0.0022\\
ConSC~\cite{cc14}&0.2781$\pm$0.0411&0.2230$\pm$0.0148&0.2139$\pm$0.0078&0.2106$\pm$0.0058&0.2884$\pm$0.0896\\
PVC~\cite{cc27}&0.5015$\pm$0.0438&0.5424$\pm$0.0537&0.6277$\pm$0.0402&0.6833$\pm$0.0931&0.7546$\pm$0.1091\\
IMG~\cite{cc28}	&0.4373$\pm$0.0100&0.4508$\pm$0.0254&0.4868$\pm$0.0147&0.5055$\pm$0.0131&0.5176$\pm$0.0415\\
PVC-GAN~\cite{partialWANG18}&0.5210$\pm$0.0090&0.6711$\pm$0.0107&0.8631$\pm$0.0043&0.9154$\pm$ 0.0107&0.9498$\pm$0.0026\\
  \textbf{GP-MVC}&\textbf{0.5874$\pm$0.0249}&	\textbf{0.7868$\pm$0.0234}&	\textbf{0.8879$\pm$0.0128}&	\textbf{0.9319$\pm$0.0082}&	\textbf{0.9655$\pm$0.0088}\\
\bottomrule
\end{tabular}}
\label{tab1}\vspace{-0.4cm}
\end{center}
\end{table*}

\begin{table*}[t]
\begin{center}
\caption{The average Clustering Accuracy in terms of different impartial ratios on the sampled MNIST Database.}
\scalebox{1}{
\begin{tabular}{ccccccc}
  \toprule
{\textbf{Methods}}&\textbf{0.1}&{\textbf{0.3}}&\textbf{0.5}& {\textbf{0.7}}& {\textbf{0.9}}\\
\midrule
best SC&0.3483$\pm$0.0080&0.3956$\pm$0.0076&0.4429$\pm$0.0114&0.4774$\pm$0.0103&0.5277$\pm$0.0106\\
AMGL~\cite{ref13}&0.1558$\pm$0.0155&0.1412$\pm$0.0218&0.1524$\pm$0.0343&0.2415$\pm$0.0631&0.3346$\pm$0.0288\\
RMSC~\cite{cc17}&0.3492$\pm$0.0077&0.4150$\pm$0.0294&0.4575$\pm$0.0233&0.4960$\pm$0.0174&0.5144$\pm$0.0204\\
ConSC~\cite{cc14}&0.3704$\pm$0.0275&0.3581$\pm$0.0231&0.3674$\pm$0.0131&0.4137$\pm$0.0396&0.5088$\pm$0.0299\\
PVC~\cite{cc27}&0.3525$\pm$0.0238&0.3864$\pm$0.0104&0.4238$\pm$0.0446&0.4401$\pm$0.0150&0.4644$\pm$0.0423\\
IMG~\cite{cc28}	&0.4655$\pm$0.0186&0.4640$\pm$0.0213&0.4613$\pm$0.0146&0.4592$\pm$0.0146&0.4622$\pm$0.0151\\
PVC-GAN~\cite{partialWANG18}&0.4517$\pm$0.0086&0.4836$\pm$0.0071&0.5280$\pm$0.0078&0.5202$\pm$0.0070&0.5340$\pm$0.0073\\
\textbf{GP-MVC}&\textbf{0.5646$\pm$0.0247}	&\textbf{0.5542$\pm$0.0330}&\textbf{0.5776$\pm$0.0169}&	\textbf{0.5963$\pm$0.0088}&	\textbf{0.5955$\pm$0.0163}\\
\bottomrule
\end{tabular}}
\label{tab2}\vspace{-0.4cm}
\end{center}
\end{table*}

\begin{table*}[t]
\begin{center}
\caption{The average Clustering Accuracy in terms of different impartial ratios on the HW Database.}
\scalebox{1}{
\begin{tabular}{ccccccc}
  \toprule
{\textbf{Methods}}&\textbf{0.1}&{\textbf{0.3}}&\textbf{0.5}& {\textbf{0.7}}& {\textbf{0.9}}\\
\midrule
best SC&0.4863$\pm$0.0122&0.5188$\pm$0.0112&0.5664$\pm$0.0143&0.6114$\pm$0.0189&0.6613$\pm$0.0178\\
AMGL~\cite{ref13}&0.6056$\pm$0.0489&0.6828$\pm$0.0564&0.7370$\pm$0.0281&0.7506$\pm$0.0320&0.7594$\pm$0.0211\\
RMSC~\cite{cc17}&0.4642$\pm$0.0159&0.5293$\pm$0.0096&0.5925$\pm$0.0154&0.6507$\pm$0.0202&0.7154$\pm$0.0375\\
ConSC~\cite{cc14}&0.5063$\pm$0.0325&0.5438$\pm$0.0272&0.5982$\pm$0.0246&0.6982$\pm$0.0481&0.7916$\pm$0.0299\\
PVC~\cite{cc27}&0.3238$\pm$0.0087 &0.3077$\pm$0.0078 &0.3419$\pm$0.0148&0.4236$\pm$0.0168&0.5730$\pm$0.0261\\
IMG~\cite{cc28}	&0.5350$\pm$0.0192 &0.5455$\pm$0.0262&0.5457$\pm$0.0193&0.5529$\pm$0.0166&0.5633$\pm$0.0213\\
PVC-GAN~\cite{partialWANG18}&0.6982$\pm$0.0104 &0.8380$\pm$0.0144&0.8806$\pm$0.0081&0.9030$\pm$0.0074&0.9234$\pm$0.0055\\
\textbf{GP-MVC}&\textbf{0.7629$\pm$0.0276}&\textbf{0.9141$\pm$0.0040}&	\textbf{0.9372$\pm$0.0056}	&\textbf{0.9454$\pm$0.0051}&	\textbf{0.9508$\pm$0.0026}\\
\bottomrule
\end{tabular}}
\label{tab3}\vspace{-0.3cm}
\end{center}
\end{table*}

\begin{table*}[t]
\begin{center}
\caption{The average Clustering Accuracy in terms of different impartial ratios on the NUS Database.}
\scalebox{1}{
\begin{tabular}{ccccccc}
  \toprule
{\textbf{Methods}}&\textbf{0.1}&{\textbf{0.3}}&\textbf{0.5}& {\textbf{0.7}}& {\textbf{0.9}}\\
\midrule
best SC&0.1863$\pm$0.0051	&0.1966$\pm$0.0070	&0.2081$\pm$0.0033&	0.2177$\pm$0.0052&	0.2263$\pm$0.0029\\
AMGL~\cite{ref13}&0.1677$\pm$0.0083&	0.1817$\pm$0.0083&	0.1850$\pm$0.0123&	0.1808$\pm$0.0076	&0.1815$\pm$0.0064\\
RMSC~\cite{cc17}&0.1925$\pm$0.0051	&0.2001$\pm$0.0084&	0.2136$\pm$0.0097&	0.2239$\pm$0.0050&	0.2287$\pm$0.0035\\
ConSC~\cite{cc14}&0.1573$\pm$0.0050	&0.1650$\pm$0.0048&	0.1804$\pm$0.0066&	0.1960$\pm$0.0058&	0.2148$\pm$0.0039\\
PVC~\cite{cc27}&0.1118$\pm$0.0017	&0.1202$\pm$0.0018	&0.1290$\pm$0.0037	&0.1482$\pm$0.0084	&0.1714$\pm$0.0114\\
IMG~\cite{cc28}	&0.1136$\pm$0.0019	&0.1215$\pm$0.0036	&0.1262$\pm$0.0028	&0.1310$\pm$0.0020	&0.1353$\pm$0.0017\\
PVC-GAN~\cite{partialWANG18}&0.1711$\pm$0.0044&	0.1988$\pm$0.0063&	0.2191$\pm$0.0050&0.2216$\pm$0.0112&0.2313$\pm$0.0082\\
\textbf{GP-MVC}&\textbf{0.1915$\pm$0.0068}&\textbf{0.2216$\pm$0.0092}	&\textbf{0.2398$\pm$0.0063}	&\textbf{0.2475$\pm$0.0046}	&\textbf{0.2770$\pm$0.0089}\\
\bottomrule
\end{tabular}}
\label{tab4}\vspace{-0.3cm}
\end{center}
\end{table*}

\begin{figure*}[t]
\centering
\includegraphics[width=17cm]{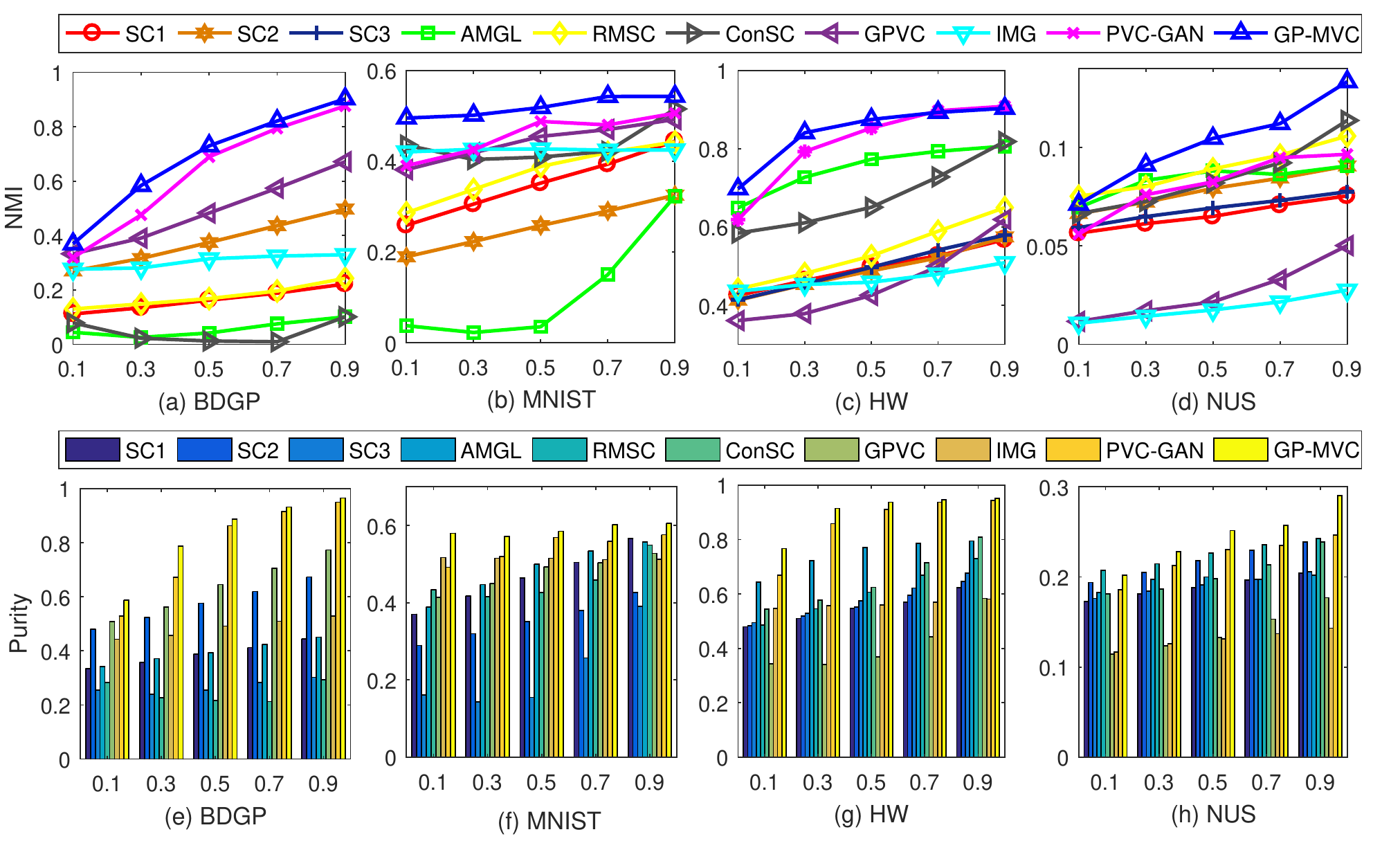}
 \vspace{-0.3cm}
\caption{The average clustering NMI and Purity of all the methods in terms of different impartial ratios on the four Databases.}\label{fig5}
 \vspace{-0.3cm}
\end{figure*}

\section{Experimental Analysis}

To test the performance of our method, we conduct several experiments on four multi-view databases and compare our method with the state-of-the-art PMVC and MVC methods.

\subsection{Experimental Setting}

\subsubsection{Dataset} We evaluate our method on four different datasets. In the following part, we present a brief introduction to these four datasets.

\textbf{BDGP~\cite{cai2012joint}:} a database that consists of both visual view and textual view. It contains $2,500$ images about drosophila embryos from $5$ categories, and each image is described by two vectors, i.e., a $1,750$-D visual vector and a $79$-D textual feature vector. In the experiment, all the data are used to evaluate the performance of the aforementioned methods on both the two features.

\textbf{MNIST~\cite{lecun1998mnist}:} a handwritten digits image database composed of $60,000$ training examples and $10,000$ testing examples, and each of them has the size of $28 \times 28$ pixels. We use the original black and white image of MNIST and its corresponding edge image for testing. Since it is difficult to conduct comparison on large-scale database, we construct a sampled MNIST database by sampling $4000$ images from the original database randomly, and then employ the sampled MNIST database to conduct the experiments.

\textbf{Handwritten numerals (HW)~\cite{van1998handwritten}:} an image database with $2,000$ images of $10$ classes from $0$ to $9$ digit. Each class contains $200$ samples with $6$ kinds of features, i.e., 76 Fourier coefficients for two-dimensional shape descriptors(FOU), 216 profile correlations (FAC), 64 Karhunen-Loeve coefficients (KAR), 240 pixel feature (PIX) obtained by dividing the image of 30$\times$48 pixels into 240 tiles of 2$\times$3 pixels and counting the average number of object pixels in each tile, 47 rotational invariant Zernike moment (ZER), and 6 morphological (MOR) features. 
In our experiment, we choose the first three views of HW database: 76 Fourier coefficients for two-dimensional shape descriptors(FOU), 216 profile correlations (FAC), 64 Karhunen-Loeve coefficients (KAR).

\textbf{NUS-WIDE (NUS)~\cite{chua2009nus}:} database consists of 269,648 images of 81 concepts. In our experiments, we select 12 categories of animal concepts, including cat, cow, dog, elk, hawk, horse, lion, squirrel, tiger, whales, wolf, and zebra. We extract three kinds of low-level features from this database: 144 color correlogram, 128 wavelet texture, 225 block-wise color moment.


\subsubsection{Baseline Methods and Evaluate Metrics}

We implement some state-of-the-art partial multi-view clustering methods. In details, we compare the proposed GP-MVC with Incomplete Multi-Modal Visual Data Grouping (IMG)~\cite{cc28}, Partial Multi-View Clustering using Graph Regularized NMF (PVC)~\cite{cc27,cc25}, Partial Multi-View Clustering via Consistent GAN (PVC-GAN)~\cite{partialWANG18}. We also compare GP-MVC with several multi-view clustering methods: Feature Concatenation Spectral Clustering (ConSC)~\cite{cc14}, Robust Multi-view Spectral Clustering (RMSC)~\cite{cc17}, Auto-weighted Multiple Graph Learning (AMGL)~\cite{ref13}, and spectral clustering of the best single view (best SC) as well.

For the partial view setting, we test the aforementioned methods under different impartial ratios. The impartial ratio is defined as the proportion of complete samples in all the samples. It varies from $0.1$ to $0.9$ with an interval of $0.2$. For MVC methods, which cannot handle missing instances, we fill in the missing instances with the average feature vector. We adopt three standard clustering validation metrics, i.e. Accuracy (ACC)~\cite{ref11}, Normalized Mutual Information (NMI)~\cite{ref24}, and Purity~\cite{ref25}, to evaluate the performance of each method.

Specifically, we randomly choose five groups of samples as missing data according to five different impartial ratios ($0.1, 0.3, 0.5, 0.7, 0.9$) in each database. In the experiment, we repeat this process 10 times. We report the average clustering accuracy and the corresponding standard deviation of all the methods on four databases, and provide the average Accuracy results in Table \ref{tab1} to \ref{tab4} respectively. Fig. \ref{fig5} shows the average clustering NMI and Purity in terms of different impartial ratios on the four databases.

\subsubsection{Implementation Details}


Our algorithm is implemented with the public toolbox of PyTorch on a desktop with Ubuntu $16.04$ system and NVIDIA Titan V Graphics Processing Units (GPUs) asa well as $32$ GB memory. We train our model using Adam~\cite{kingma2014adam} optimizer with default parameter setting, and the learning rate is fixed as $0.0001$. For each training step, we conduct $20$ epochs and record the experimental results. Besides, we also test all the performance of the other methods by Matlab on the same environment for comparison.

\begin{figure*}[!t]
\centering
\subfigure[0.1]
{ \includegraphics[width=5.8cm]{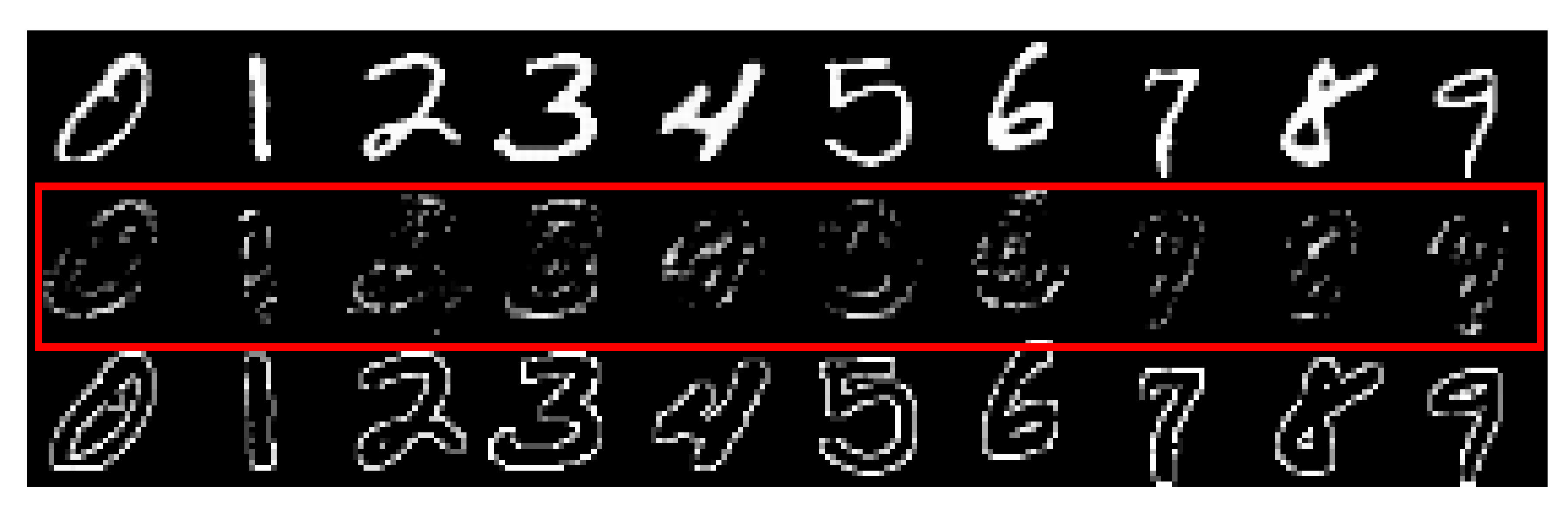}}
\subfigure[0.5]
{ \includegraphics[width=5.8cm]{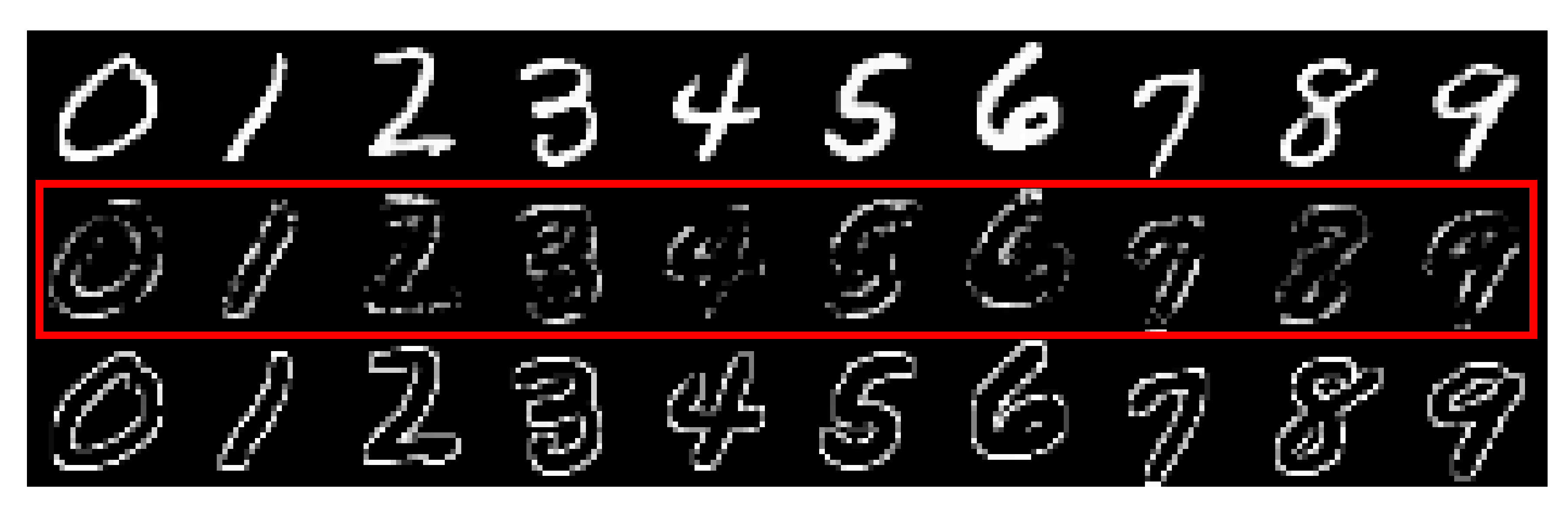}}
\subfigure[0.9]
{ \includegraphics[width=5.8cm]{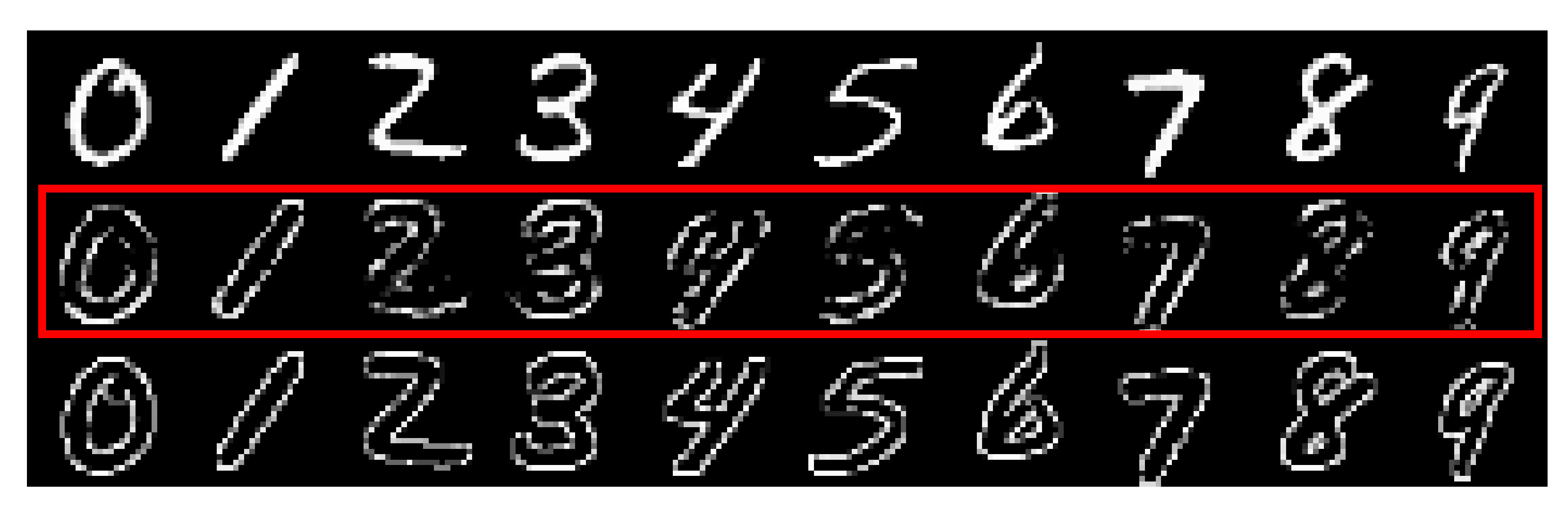}}
\subfigure[0.1]
{ \includegraphics[width=5.8cm]{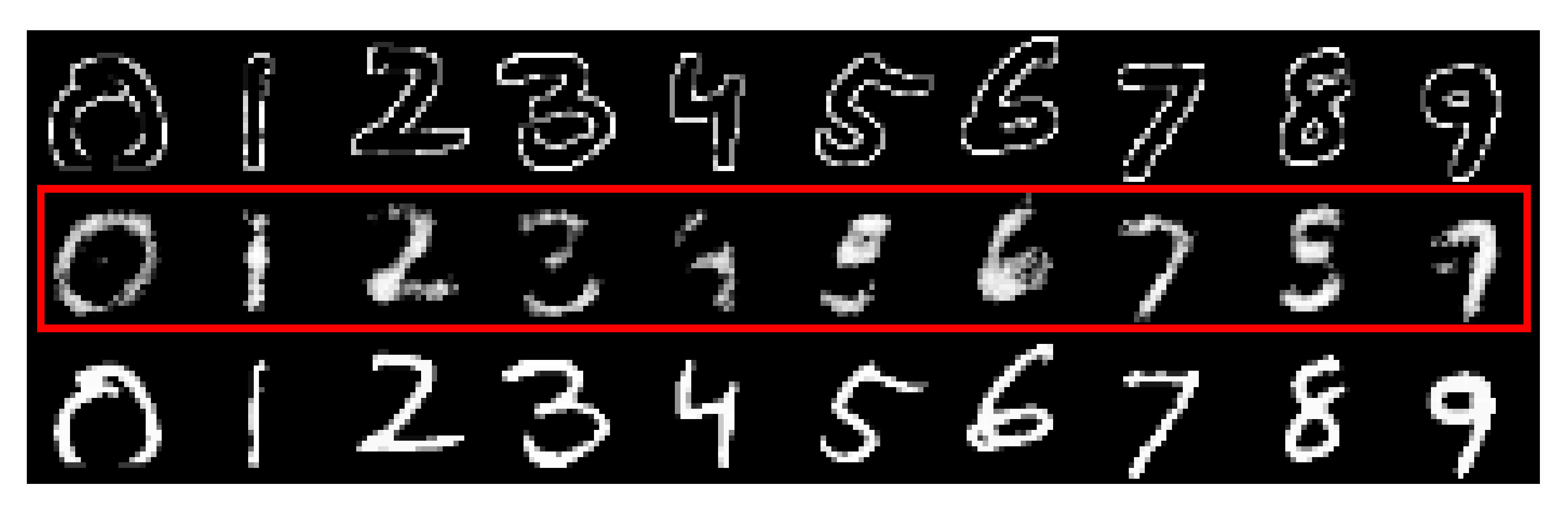}}
\subfigure[0.5]
{ \includegraphics[width=5.8cm]{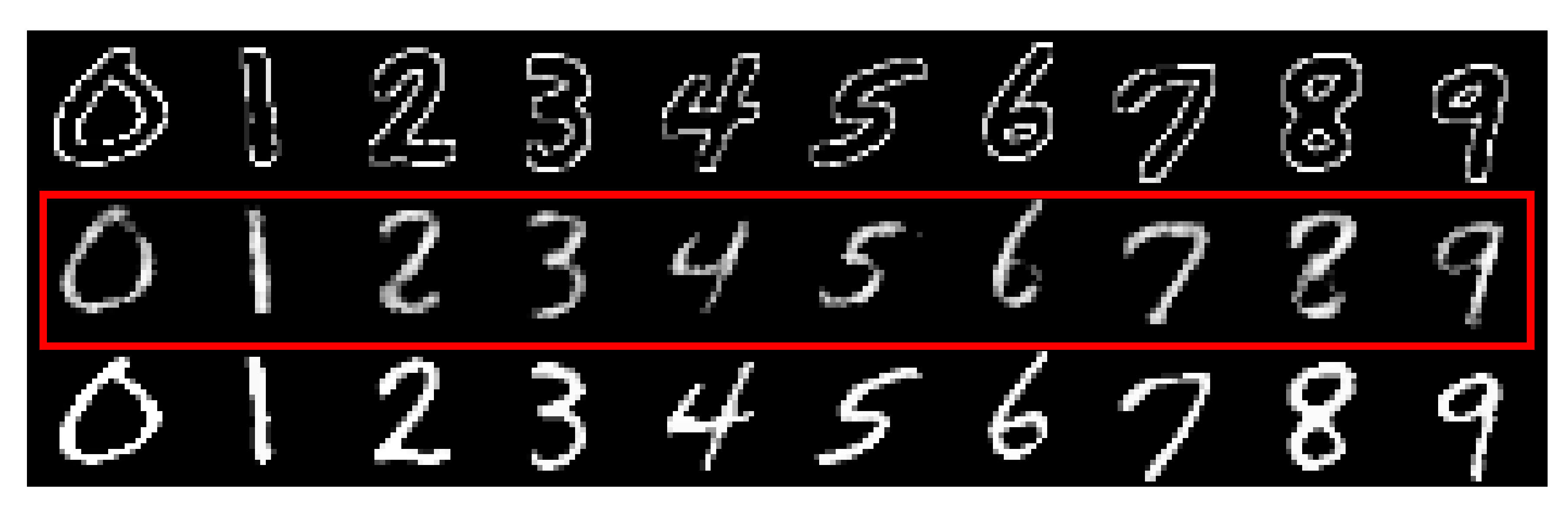}}
\subfigure[0.9]
{ \includegraphics[width=5.8cm]{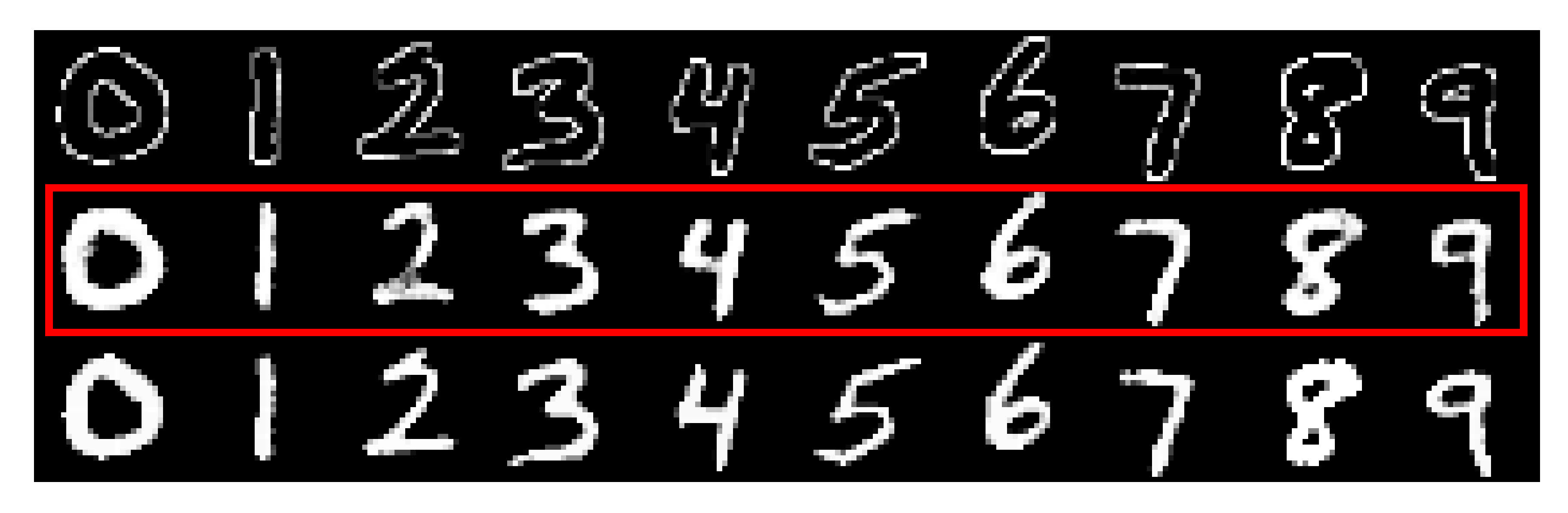}}
\vspace{-3mm}
\caption{The images in the first line are the real images from view 1 ((a), (b), (c)) or view 2 ((d), (e), (f)) of MNIST database. The images in the second line (marked by red box) are the fake images which are generated by the latent representation of the images in the first line. The images in the third line are the real images of view 2 or view 1 corresponding to the first line. The images generated under different training data corresponding to different impartial ratios. (a), (d) 0.1 impartial ratio; (b), (e) 0.5 impartial ratio; (c), (f) 0.9 impartial ratio.}\label{fig7}
\end{figure*}

\subsection{Partial Multi-view Clustering Performance}
The evaluation results are summarized in Table~\ref{tab1} to Table~\ref{tab4} and Fig.~\ref{fig5}, which indicate that our methods achieves better clustering performance than the others on all the cases. Here, we present some important observations as below:

From Table \ref{tab1} to Table \ref{tab4}, and Fig. \ref{fig5}, we could see that PMVC methods are superior in most databases, especially when the partial ratio is large. It indicates that missing-view data have a negative influence on effectiveness of MVC methods. Therefore, PMVC methods are more effective when confronting missing data problem. From Table \ref{tab2}, we could see that when the view is not complete, the multi-view clustering method AMGL has inferior performance than single-view methods. It further illustrates that some of multi-view methods are sensitive to missing data or noises.

The results of Table \ref{tab1} to Table \ref{tab4} also demonstrate that our method outperforms methods tested. It is probably because of that GP-MVC is able to learn a consistent clustering structure for each view, with which it effectively generates the missing data. Thereby, it can construct a more effective common subspace with these complementary missing data. Compared the results of PVC-GAN and GP-MVC in Table \ref{tab1} and Table \ref{tab2}, we can observe GP-MVC perform better than PVC-GAN. The difference of these two only is the fusion way. In GP-MVC, we use weighted adaptive fusion way to fuse the multiple latent space. It prove the effectiveness of the weighted adaptive fusion loss.

From Table \ref{tab3} to Table \ref{tab4}, we can see that partial multi-view methods (PVC, IMG, and PVC-GAN) perform worse than others. It is because all these methods can only be conducted on the two-view databases. While HW and NUS have more than two views. Thus, for PVC, IMG, and PVC-GAN, we use the first two views of HW and NUS. For the MVC methods, we still use the average sample to fill up the missing sample. The experimental results illustrate our model can well be applied to multi-view databases.


\begin{table}[t]
\begin{center}
\caption{The ablation study of GP-MVC under different impartial ratios on the HW dataset. We show the clustering accuracy of our method with different loss function.}
\scalebox{1}{
\begin{tabular}{ccccccc}
  \toprule
{\textbf{Loss}}&\textbf{0.1}&{\textbf{0.3}}&\textbf{0.5}& {\textbf{0.7}}& {\textbf{0.9}}\\
\midrule
AE&0.6530&0.7910&0.8760&0.8835&0.8945\\
AE + AT&0.7165&0.8680&0.8840&0.8985&0.9000\\
ALL&\textbf{0.7629}&\textbf{0.9141}&\textbf{0.9372}&\textbf{0.9454}&\textbf{0.9508}\\
\bottomrule
\end{tabular}}
\label{tab5}\vspace{-0.3cm}
\end{center}
\end{table}

\begin{table*}[t]
\begin{center}
\caption{The Clustering performance of our method on the whole MNIST Database.}
\scalebox{1}{
\begin{tabular}{ccccccc}
  \toprule
{\textbf{Clustering metrics}}&\textbf{0.1}&{\textbf{0.3}}&\textbf{0.5}& {\textbf{0.7}}& {\textbf{0.9}}\\
\midrule
Accuracy&0.5016$\pm$0.0074&0.5144$\pm$0.0070&0.5203$\pm$0.0097&0.5391$\pm$0.0104&0.5551$\pm$0.0120\\
NMI&0.4567$\pm$0.0018&0.4645$\pm$0.0047&0.4659$\pm$0.0045&0.5143$\pm$0.0029&0.4828$\pm$0.0167\\
Purity&0.5398$\pm$0.0033&0.5555$\pm$0.0043&0.5567$\pm$0.0073&0.5689$\pm$0.0049&0.5757$\pm$0.0049\\
\bottomrule
\end{tabular}}
\label{tab6}\vspace{-0.3cm}
\end{center}
\end{table*}

\subsection{Model Discussion}

\subsubsection{Ablation Study}
To verify how much effect of each term in the objective function of our model, we conduct ablation studies. We perform the following three experiments to isolate the effect of the loss ${L_{\rm AE}}$, ${L_{\rm AT}}$, ${L_{\rm FU}}$ and ${L_{\rm KL}}$. In the first experiment, we only use auto-encoder loss ${L_{\rm AE}}$ to train encoder and generator networks. In the second experiment, we use auto-encoder loss ${L_{\rm AE}}$ and adversarial training loss ${L_{\rm AT}}$ to train encoder, generator and discriminator networks. In the third experiment, we use all objective function to train our model, i.e., based on the second experiment, we add adaptive fusion loss ${L_{\rm FU}}$ and KL-clustering loss ${L_{\rm KL}}$. In each experiment, we change the partial ratio from $0.1$ to $0.9$ with an interval of $0.2$, and do clustering task on the common representation ${\bf{Z}}$ learned by encoder network. Then we show the clustering accuracy on Table \ref{tab5}. From Table \ref{tab5}, we can see that the third experiment using all objective function has the best performance, and the clustering accuracy of using auto-encoder loss ${L_{\rm AE}}$ and adversarial training loss ${L_{\rm AT}}$ is superior to that only using auto-encoder loss. It illustrates each term in our objective function has great effect on the final clustering performance of our model. When adding adversarial training loss ${L_{\rm AD}}$ to auto-encoder loss ${L_{\rm AE}}$, the clustering accuracy is improved. It is probably because that generated sample for missing view promote to learn a better clustering structure and improve clustering performance. Weighted adaptive fusion loss ${L_{\rm FU}}$ and KL-clustering loss ${L_{\rm KL}}$ boost the performance after adding it in the third experiment. This phenomenon illustrates that a good common representation will in turn helps generate more realistic samples for missing view and improve clustering performance again.

\subsubsection{Missing Data Generation}
To visually present the generated results of our method, we show the generated images (marked by red box) by our method on the sampled MNIST database under three different impartial ratios (0.1, 0.5, 0.9) on Fig \ref{fig7}. We can see our method can well generate the missing data. With the number of paired data increasing, our method can generator more real data.

\subsubsection{Large-Scale Partial Multi-View Clustering}
In addition, we conduct all methods on the total MNIST database in terms of different impartial ratios and run 10 times. Due to all the other methods go wrong with the problem of out of memory. We only show the clustering performance of our method in Table \ref{tab6}. All other methods cannot clustering on $60000\times2$ images as for the memory of computer does not satisfy the need of other algorithm. This illustrate our method can be used on large scale database.

\section{Conclusions}
In this paper, we propose a novel generative partial multi-view clustering approach. It is able to fill up the incomplete views based on the common subspace via GAN model, and learn an excellent clustering structure simultaneously. In addition, it further employs the complementary incomplete view to study a consistent common structure and greatly improves its clustering performance. We validate the clustering performance improvement of the proposed method via a series of comprehensive experiments, and comparison results to several existing methods demonstrates the superiority of GP-MVC.


%





\ifCLASSOPTIONcaptionsoff
  \newpage

\fi
\small{
\bibliographystyle{IEEEtran}
\bibliography{egbib}
}

\end{document}